\def\year{2020}\relax
\documentclass[letterpaper]{article} % DO NOT CHANGE THIS
\usepackage{aaai20}  % DO NOT CHANGE THIS
\usepackage{times}  % DO NOT CHANGE THIS
\usepackage{helvet} % DO NOT CHANGE THIS
\usepackage{courier}  % DO NOT CHANGE THIS
\usepackage[hyphens]{url}  % DO NOT CHANGE THIS
\usepackage{graphicx} % DO NOT CHANGE THIS
\usepackage{graphicx}  % DO NOT CHANGE THIS
\usepackage[linesnumbered,ruled]{algorithm2e}
\newcommand{\citet}[1]{\citeauthor{#1}~\shortcite{#1}}
\newcommand{\citep}{\cite}

\title{Predicting Plans and Actions in Two-Player Repeated Games}%\\
\author{Najma Mathema, Michael A. Goodrich, and Jacob W. Crandall\\
Computer Science Department\\
Brigham Young University\\
Provo, UT, USA\\
}%\\

\begin{document}

\maketitle

\begin{abstract}
Artificial intelligence (AI) agents will need to interact with both other AI agents and humans.
%One way to enable effective interaction is to c
Creating models of associates help to predict the modeled agents’ actions, plans, and intentions. 
%If AI agents are able to predict what other agents in their environment will be doing in the future and can understand their intentions, they can use these predictions in their planning and decision-making. 
This work introduces algorithms that predict actions, plans and intentions in repeated play games, with providing an exploration of algorithms.
%Prior work~\cite{crandall2018cooperating,crandall2014towards} introduced the S\# algorithm, which is designed as a robust algorithm for many two-player repeated games. The proposed work is to create models for predicting actions, plans, and intents of an S\# agent using observations of its actions and speech acts as well as the actions and speech acts of the other agent playing the game. 
%Because S\# generates actions, has (internal) experts that seek to accomplish an internal intent, and associates plans with each expert, it is a useful algorithm for exploring intent, plan, and action in repeated games.
%najma edit-commented a few lines and added intro of S#
We form a generative Bayesian approach to model S\#. 
%and utilize the past observations to estimate the internal states and state transitions used by S\#. S\# 
S\# is designed as a robust algorithm that learns to cooperate with its associate in 2 by 2 matrix games. The actions, plans and intentions associated with each S\# expert are identified from the literature, grouping the S\# experts accordingly, and thus predicting actions, plans, and intentions based on their state probabilities. Two prediction methods are explored for Prisoners Dilemma: the Maximum A Posteriori (MAP) and an Aggregation approach. 
%The action predictions were compared with other strategies presented in \cite{fudenberg2012slow}. 
MAP ($\approx$ 89\% accuracy) performed the best for action prediction.
%while Tit-for-tat's performance was comparable to that of MAP. 
Both methods predicted plans of S\# with $\approx$ 88\% accuracy. 
%Paired T-test shows that the difference in the average performance between the MAP and Aggregation for predicting S\#'s actions is not big enough to be statistically significant. However, when predicting actions in absence of speech acts, T-test shows that MAP performs significantly better than Aggregation. 
Paired T-test shows that MAP performs significantly better than Aggregation for predicting S\#'s actions without cheap talk. Intention is explored based on the goals of the S\# experts; results show that goals are predicted precisely when modeling S\#. The obtained results show that the proposed Bayesian approach is well suited for modeling agents in two-player repeated games.
%najma edit-end
%Intent is studied in terms of the desired outcome, tactics and plan. 
%accumulating probability across experts. 

\end{abstract}

\noindent When agents interact, it is useful for one agent to have an idea of what other agents are going to do, what their plans are, and what intentions guide both their plans and actions. 
%It is important that agents should be able to interact effectively for long-term and should also be explainable(be able to explain its own and others’ actions). 
%This requires the agents to have the ability to deduce the intentions, attitudes, plans, strategies, behaviors, expectations of other agents. 
%This is the goal of our research,i.e. to be able to 
This work creates agent models that allow utilizing the observations from their past interactions to predict the modeled agent's actions, plans, and intentions to  develop algorithms that: (a)~predict actions of another agent and (b)~identify their plans and intent.

The main concept of the work is based on the perspective from the literature in which \textit{intentions} are the reasons behind \textit{actions}, and that \textit{plans} are the means for mapping intentions to actions~\cite{perner1991understanding,bratman1987intention,malle1997folk}. The overarching objective of the research is to infer another agent's plans, goals, intentions and predict behavior (actions) based on these inferences and using observations of their actions in an environment. If an AI agent is able to predict future actions, the agent can plan ahead for appropriate actions and hence be able to make better decisions for the future. 

This work makes predictions in the context of Repeated Games (RGs). Game theory has been applied in numerous ways to understand human/agent behavior, relationships, and decision-making. RGs in game theory provide an environment for understanding and studying the relationship between agents because the game construct requires each agent to account for the consequence of its action on the future action of the other agent. The dilemma of whether to cooperate or to compete with each other has been extensively studied in the game Prisoners Dilemma in the literature of psychology, economics, politics and many other disciplines. Hence, the same game has been used for this study. 
%This provided the motivation to study the interaction in the scenario of Repeated Games which can later be utilized and applied in real life scenarios.
%najma edit-added an intro about S#
Prior work~\cite{crandall2018cooperating,crandall2014towards} introduced the S\# algorithm, which is designed as a robust algorithm that learns to cooperate with its associate in many 2 by 2 matrix games.
%which is designed as a robust algorithm for all two-player repeated games.
S\# is built on top of S++ \cite{crandall2014towards} with the ability to share costless, non-binding signals called ``cheap talk" indicating its experts' intentionality. For better expert-selection, in each round prior to taking an action, the players get an opportunity to communicate by sharing their plans via cheap talk. This paper presents a model for predicting actions, plans, and intents assuming the agent to be modeled is an S\# agent. S\# is studied because it is a highly effective algorithm in RGs and it uses explicit models of planners (called ``experts") that are motivated by specific designer intentions~\cite{crandall2018cooperating,oudah2018ai}.
%najma edit- end
In the context of modeling S\#'s behavior in RGs, we use a generative Bayesian model, which assumes that agents have a number of internal states defining the ``state of mind'' used to select what action they would want to take given the observations they see. The observations are (a)~the speech acts/proposals via cheap talk that the players share with each other prior to taking their action and (b)~the actions taken by both the S\# agent and the agent with whom S\# is interacting. Table~\ref{tab:interaction} shows a few interactions of S\# against a human player ABL in Prisoners Dilemma.

\begin{table}[htb]
    \begin{tabular}{ |p{.85cm}|p{.85cm} | p{3.2cm}| p{.89cm}|p{.89cm} | }
     \hline
     \textbf{Round} & \textbf{Player} & \textbf{Speech acts} & \textbf{Action}& \textbf{Payoff} \\ 
     \hline\hline
     35 & S\# & None & B & 20\\
     \hline
     35 & ABL & You betrayed me. Curse you. & D & 20\\
     \hline
     36 & S\# & None  & B & 100\\
     \hline
     36 & ABL & None & C & 0\\
          \hline
     37 & S\# & In your face! I forgive you. Let’s alternate between AC and BC. This round, let’s play BC. Do as I say or I will punish you.  & B & 100\\
     \hline
     37 & ABL & Let’s always play BD. & C & 0\\
          \hline
     38 & S\# & Excellent. This round, let’s play AC.  & A & 60\\
     \hline
     38 & ABL & None & C & 60\\
          \hline
     %39 & S\# & This round, let’s play BC.  & B & 100\\
     %\hline
     %39 & ABL & Excellent. Sweet. We are getting rich. & C & 0\\
     %\hline
    \end{tabular}
\caption{S\# vs human player ABL in the Prisoner’s Dilemma}
\label{tab:interaction}
\end{table}

The generative Bayesian model provides a probability distribution over the S\# agent's internal states.  This probability distribution can be used as input to algorithms that predict the most likely state, the most likely action, the most likely plan, and the most likely intention. Additionally, this kind of Bayesian model is not dependent on the type of RGs, and thus could be used for many two-player RGs. Since the model is based on observations in the environment and employs Bayesian reasoning, it does not require a huge dataset to train the model for better performance. Two types of algorithms for translating the distribution into predictions will be explored: (a) a Maximum A Posteriori (MAP) estimate of the most likely state, which implicitly identifies an action, plan, and intention; and (b) estimates of actions, plans, and intentions that aggregate probability over related states. 
%najma edit- commented out motivation section
\iffalse
This work is motivated by three broad problems: The first motivation is that using ideas from the folk psychology, namely the intentions that frame plans and actions, as a means for creating models of other agents might lead to explainable AI algorithms. The second motivation is that by learning to model other agents, we can learn lessons about modeling that allow an AI agent to be able to assess its own proficiency at a task. Each of these first two motivations is discussed in the review of related literature. 

The third broad motivation is that modeling an S\# agent builds the groundwork for modeling other types of agents. S\# makes actions, hidden states, and plans explicitly while allowing different implicit intentions encoded by the human designer through his or her choice of expert. Thus, modeling S\# agents can help us learn how to model other game-playing agents like humans. 
\fi
%najma edit-end
\section{Related Literature}
%Repeated Games~(RGs) are simply single stage games being played repeatedly, either a fixed number of rounds or a random number of rounds unknown before play. %The proposed work considers two player games, with players denoted by \textit{player}$i$ and \textit{player}$\neg i$. 
A variant of RGs implemented by  Crandall et al.~ \cite{crandall2018cooperating,oudah2018ai}, called RGs with ``cheap talk'', allows each player to share messages/ proposals with the other player before actions are taken.%Cheap talk gives player~$i$ an opportunity to share its plans, sentiments, threats, bluffs, etc., in an effort to influence the \textit{player}$\neg i$ in a way that facilitates the \textit{player}$_i$'s goal. Players can be humans or artificial agents, enabling human-human, human-agent, or agent-agent interaction. 

%S\# is an algorithm developed by \cite{crandall2018cooperating,oudah2018ai} for effectively cooperating with agents to maximize payoffs while minimizing disappointment. S\# is built on top of S++, sharing the same internal states, but with the ability of cheap talk to share the intentionality of the players. S++ has been tested extensively in agent-agent interactions, and has been evaluated in human-agent interactions \cite{crandall2018cooperating,crandall2014towards,oudah2018ai}.

\subsection{Intentional Action}
Consider the motivation of using the notion of ``intentionality" from folk psychology as the basis for modeling other agents. As per~\cite{de2017people}, people regard ``Autonomous Intelligent Systems" as intentional agents; people, therefore, use the conceptual and psychological means of human behavior explanation to understand and interact with them. Folk Psychology suggests that it is the belief, desire, and intention of humans that control human behavior and that our intention is the precursor to the action we take~\cite{perner1991understanding}. Hence, inferring the intent behind a particular action allows a human to infer the plans and goals of the agent. In this context, \textit{intent} is associated with the commitment to achieve a particular goal through a plan~\cite{bratman1987intention}. %``Commitment" is the key, as it allows prioritizing resources to be used in pursuing the intent rather than other desires. For this work, we propose an operational definition of intent as the process of forming plans to achieve goals where the goals may also be a subset of other goals, and the plans are the actions one would take to achieve those goals. 
Once an intent is formed and a plan is selected to achieve the desire, an ``intentional" action is one that is derived as an instrumental means for moving towards the intent.

M.~de Graaf et al. mention that so-called ``Autonomous Intelligent Systems'' exhibit traits like planning and decision making, and hence are considered ``intentional agents''~ \cite{de2017people}. They further claim that the behaviors of intentional agents can be explained using the human conceptual framework known as \textit{behavior explanation}.

Related to the literature on intentional agents is work in ``folk psychology''~\cite{perner1991understanding,bratman1987intention,malle1997folk}, in which agent \textit{beliefs}, \textit{desires}, and \textit{intentions} are used to explain how and why agents choose actions towards reaching their goal. 
%The main reason behind performing any action is considered to be the \textit{intention} behind that action, though there are different factors that come before and in between mapping intentions to actions~\cite{bratman1987intention,malle1997folk,fishbein2011predicting,ajzen1991theory}. 
%najma edit-cited the suggested paper
%najma left
Baker et al. (2011)~\nocite{baker2011bayesian} presented a Bayesian model of human Theory of Mind based on inverse planning to make joint inferences about the agents’ desires and beliefs about unobserved aspects of the environment. Similar to our work, they model the world as a kind of Markov Decision Process and use the observations in the environment to generate posterior probabilities about the environment. Additionally, by inverting the model, they make inferences about the agents’ beliefs and desires. 
%najma edit-end
%Complementing and extending the folk psychology work on intentions, \textit{The Theory of Reasoned Action and Planned Behavior} provides another model for behavior explanation~\cite{fishbein1977belief,fishbein2011predicting,ajzen1991theory}. This model explains actions based on attitudes, intentions, subjective norms, perceived behavioral control (PBC), and behavioral, normative and control beliefs.

%Hence, we think that knowing one’s current actions and also being able to tell where one will be in the future with its actions, it can also help evaluate the proficiency of the agent along with inferring plans and intentions.

\subsection{Modeling Other Agents}

Seminal work on modeling agents in the field of game theory was presented in~\cite{axelrod1981evolution}. Axelrod's models allow strategies to play against each other as agents to determine the winning strategy in Prisoners Dilemma tournaments. Early work on agent modeling tended to focus on equilibrium solutions for games and has now extended to various fields of computer science like~\cite{lasota2017survey,stone2000multiagent,kitano1997robocup}. %najma left ? reference???

One modeling approach is to predict action probabilities for the modeled agent, an early example which is the Fictitious Play algorithm~\cite{brown1951iterative}. %where the modeling agents learn probabilities by counting the number of times actions are played in simulated interactions. 
In contrast to the simple empirical probability distributions of fictitious play, other authors have worked on making action predictions by learning the action probabilities of the opponent conditioned on their own actions~\cite{sen1997learning,banerjee2007reaching}.  

%Modeling based on the so-called agent \textit{type} or behavior of the players, \cite{harsanyi1967games} mentions each player having belief about the joint probability distribution indicating the selection of the possible types of other players. In the implementation of S\#, the types or the experts have been specified manually~\cite{crandall2014towards}, whereas there are other works in literature  where agents do not have such prior knowledge of experts, and is based on interaction history~\cite{barrett2013teamwork}. Albrecht et al.~\cite{albrecht2017special} present research directed towards effective multiagent interaction without prior coordination. 

Similar to our research objective, which is to be able to predict the next moves of the opponent, \citet{gaudesi2014turan} worked on an algorithm called Turan to model the opponent player using finite state machines. The work in~\cite{deng2015study} studies Prisoners Dilemma as a game with incomplete information and using Bayes rule and past interaction history to form a possibility distribution table for each players' choice to predict the players' choices. %; this work found the Bayes Model to be superior to other strategies. Another approach that used a Bayesian model for action prediction, predicted users’ next actions in a command line shell using probabilistic model conditioned on user's previous commands~\cite{davison1998predicting}.
%najma edit-cited additional paper
%najma left cite author et al
Park et al. (2016)~\nocite{park2016active}%~\cite{park2016active} 
assert that building precise models of players in Iterated Prisoners Dilemma requires a good dataset, so they use a Bootstrap aggregation approach to generate new data randomly from the original dataset. Also, an observer uses active learning approach to model the behavior of its opponent.
%najma edit-end

\subsection{Inferring Intent}
There has not been much research in predicting or inferring intents of agents in RGs, but there has been previous work in predicting the intent of agents in various other fields. Intent in prior work relates to goals and plans. 
%najma left cite author
Kuhlman et al. (1975)~\nocite{kuhlman1975individual} talk about the goals of agents in mixed-motive games by identifying their motivational orientation (cooperative, individualistic, or competitive) based on their choice behavior in decomposed games. Thus, knowing the motive of the subject, they use it to predict actions for Prisoners Dilemma. The work in~\cite{cheng2017predicting} linked intent with goal specificity and temporal range when predicting intents in online platforms. Very recent research work uses deep-learning models for intent prediction \cite{qu2019user,pirvu2018predicting}. Rabkina et al. (2013)~\nocite{Rabkina2013analogical} used a computational model based on analogical reasoning to enable intent recognition and action prediction.
%\cite{qu2019user}~used traditional machine learning classifiers, and also compared the results to deep-learning models like Convolutional Neural Netwok (CNN) and bi-directional for labeling the information-seeking conversations (MSDialog data) ~\cite{bhatia2012classifying,qu2018analyzing}. Similarly, \cite{pirvu2018predicting} used a recurrent neural network and a CNN model to predict user intent based on the queries used to access a certain webpage, where the intent group was identified as informational, transactional and navigational based on the taxonomy from~\cite{broder2002taxonomy}. 
Other methods that use Bayesian models for intent prediction include~\cite{mollaret2015perceiving,tavakkoli2007vision,rios2012intention}. %Using a multimodal perceptual system fused with an HMM that gives a prior probability estimate of the user’s intention, the work in~\cite{mollaret2015perceiving} focused on detecting user’s intention-for-interaction. Intention is detected by a thresholded output of the posterior estimate. Likewise using an HMM, work in~\cite{tavakkoli2007vision} encoded the underlying intent of actions in hidden states; the algorithm was able to detect the intention before the goal was achieved. 
%It had parameters encoding the goals of an activity change like distance, angle and the state with maximum probability is chosen to detect the underlying intent. 
%Work in~\cite{rios2012intention} treated the user intent as a goal, where the estimation of intent or the goal is done using a Bayesian framework combining the user input and prior knowledge on environment to yield a posterior probability distribution over its set of goals.

\section{Modeling Framework}
\subsection{Bayesian Graphical Model}
\begin{figure*}%[hbt]
    \centering
    \includegraphics[width=3.25in]{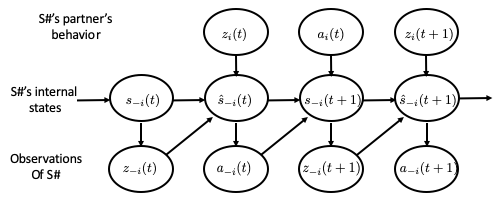}
    \caption{Modeling S\#}
    \label{fig:ModelingSSharp}
\end{figure*}

A Bayesian graphical model is used to model S\#. The model begins with a prior probability distribution over possible S\# agent states and then propagates that distribution using observations of actions and speech-acts/proposals.  

The structure of the Bayesian model is illustrated in Figure~\ref{fig:ModelingSSharp}. The model makes it evident that future predictions are based on the present state and immediate observations. Understanding this model is made easier by comparing it to a Hidden Markov Model (HMM). An HMM is a five-tuple \[{\rm HMM} = \Big(S,O,p(s_0),p(s_{t+1}|s_t),p(o_t|s_t) \Big),\] where $S$ is  a finite set of (hidden) states, $O$ is a finite set of observations, $p(s_0)$ is the initial state distribution (i.e., the distribution over states at time~$t=0$), $p(s_{t+1}|s_t)$ represents the transition probability function that describes how states change over time, and $p(o_t|s_t)$ is the emission probabilities (i.e., the probability that a given observation $o_t$ is generated by hidden state $s_t$). An HMM is one of the most simple dynamic Bayesian models because it describes how states change over time. A common application of HMMs is to try to infer a most likely hidden state from a series of observations.

Like an HMM, our model is also a dynamic model, but the inference task is slightly different and so are the model elements. The proposed model differs from the traditional HMM in two ways: First, the Bayesian model makes two state transitions at a single time step, that is, there are two hidden states at each time step. Second, there is an external input to the model. Figure~\ref{fig:ModelingSSharp} illustrates the proposed model. In the figure, the player being modeled is denoted by a subscript `-i' whereas the player's associate (in the game) is denoted by a subscript `i'. 
%The notation follows a useful mnemonic: ``I (agent~$i$) am playing the repeated game with agents who are not me (agent~$-i$)," where $-i$ is read ``not~$i$".

The Bayesian model is a tuple with seven elements, 
\begin{eqnarray*}
    {\sc BModel} &=&  \Big(S, O, \Sigma, p(\hat{s}_{-i}(t) | {s}_{-i}(t),z_{-i}(t),z_i(t)),  \\
    && p({s}_{-i}(t+1) | \hat{s}_{-i}(t),a_i(t), a_{-i}(t)),\\
    && p(z_{-i}(t)|s_{-i}(t)), p(a_{-i}(t)|\hat{s}_{-i}(t)) \Big).
\end{eqnarray*}
As with the HMM, $S$ represents the set of states and $O$ represents the set of observations. The set of states has two different kinds of states, $s_{-i}\in S$, which represent propositional states (states from which speech acts are generated) and $\hat{s}_{-i}\in S$, which represent action states (states from which game actions are chosen).
The set of observations $O$ has two different kinds of observations: (1)~the speech acts/proposals, $\{z_{-i}\}\in O$, shared by the player being modeled and (2)~the action, $\{a_{-i}\}\in O$, taken by the player being modeled.  $\Sigma$ is the set of exogenous inputs to the model, which consists of the observed actions and speech acts of the other player in the game, represented by $a_{i}$ and $z_{i}$, respectively.

As mentioned earlier, the model of agent~$-i$ is based on a time series with two types of hidden states, $s_{-i}(t)$ and $\hat{s}_{-i}(t)$. The proposed model takes two state transitions at a single time step. For a single time step, the first state transition occurs from $s_{-i}(t)$ to $\hat{s}_{-i}(t)$ based on the observation of what proposals are shared. $\hat{s}_{-i}(t)$ is a kind of temporary state for S\#, from which it generates its aspiration level to choose the expert to play the game further. The next state transition from $\hat{s}_{-i}(t)$ to $s_{-i}(t+1)$ takes place based on the observation of the actions of the players. 
%The observation proposal at a time 't' is indicated by z(t), and the observation of action at a time t is indicated by action a(t). 
This state transition gives the prediction for the state at the next time step, which is then utilized in predicting the action of the modeled player for the next time step.

$S(t)$ is the set of states available to the modeling agent~$i$ at time~$t$, and is given by the union of all the states in each expert's state machine,
\[
    S(t) = \cup_{j\in {\mathcal J}} S_{\phi_j}.
\]

\subsection{Conditional Probabilities}
The Bayesian model makes use of the priors and conditional probabilities to find the posterior probability of the states after each observation. The priors about the states of the experts represent the agent's beliefs about the states. Ideally, prior probabilities should be fairly close to the true probabilities in real scenarios; prior probabilities affect the future computations and the predictions to be made. For computing the priors, the initial knowledge about S\# and how the experts are formed and then selected is utilized. 

Based on the game, S\# generates a set of experts, which are essentially strategies that employ learning algorithm to select actions and generate and respond to speech acts, based on the state it is in. To select an expert to take an action, the ``expert potential" needs to meet a specified aspiration level and the expert also needs to carry out plans which are congruent with its partner’s last proposed plan. Thus, the priors for models of S\# agents are the probabilities that S\# selects a particular expert in the first round based on the expert’s potential and uniform distribution over the aspiration level, with most of the probability assigned to the start state of the expert.

The following conditional probability elements describe the necessary components for designing the model. %Each element is given a name that can act as a mnemonic for remembering what is encoded in the conditional probability. Also, 
The notation used in the conditional probabilities is given by:
\[ \: a : action, \: z : proposal, \: i : Partner, \: -i : S\#.\]
\begin{enumerate}

    \item Sensor Model 1 (Speech): Given the current state of the agent, the sensor model provides the probability of seeing a particular proposal.% Thus, at a time step~$t$, the state probabilities of S\# are updated based on the observation of the proposal of the player being modeled. 
    \[
    {p(z_{-i}(t) \: | \: s_{-i}(t))}
    \]

    \item Transition Model 1 (Reflection): For the same time step, the transition model is used for transitioning to a new state after the proposals are observed. 
    \[ 
    {p( \hat{s}_{-i}(t) \: | \: s_{-i}(t),z_{i}(t),z_{-i}(t)) }
    \] 
    %This is called as the reflection model.

    \item Sensor Model 2 (Action): Given the state of the agent, the sensor model provides the probability of seeing a particular action.
%    After the proposals are shared, the action model is used for taking action given the current state.
    \[ 
    {p(a_{-i}(t) \: | \: \hat{s}_{-i}(t))}
    \]

    \item Transition Model 2 (Update): Once the actions are taken by both the players, the update model encodes how a state transition occurs to a new state for the next time step. 
    \[ 
    {p(s_{-i}(t+1) \: | \: \hat{s}_{-i}(t),a_{i}(t),a_{-i}(t)) }
    \]

\end{enumerate}

The S\# algorithm has Finite State Machines (FSMs) for each of the experts which define what speech acts are to be generated based on the internal states of the experts and the events in the game~\cite{crandall2018cooperating}. The events in the game could involve the event of selecting an expert or the events that affect individual experts. These FSMs have state transition functions that map the events, internal states, and speech outputs. Hence, for the game, the above conditional probabilities have been determined based on how S\# acts in each event and also adding some uncertainty to make sure that other possible transitions are non-zero.

\subsection {Updating the Probability Distribution -- Bayes Filter Algorithm}
%\mg{Tell the reader what is in this section and why its important. Tell them how it reviews the Bayes Filter algorithm, and tell them that the algorithm is used to update the probability distribution over states given external inputs and observations.}

%The proposed Bayes Filter model is initially implemented to model S\#’s behavior in a human- S\# repeated game. Assuming that humans have the same internal states as that of S\# and have similar state experts in mind, it is then used to model a human- human repeated game. All the observations to the current time are included in the model as presented in the Figure~\ref{fig:ModelingSSharp}. The traditional Bayes Filter equations were modified as it required multiple state transitions in a single time step.

An algorithm is needed to aggregate observations into a distribution over the hidden states. Since the proposed model is a dynamic Bayesian model, a Bayes Filter is an appropriate algorithm~\cite{Thrun}. The Bayes Filter is a general algorithm for estimating an unknown probability density function over time given the observations. The beliefs are computed recursively and are updated at each time step with the most recent observations. The algorithm presented in Algorithm \ref{alg:Bayes} is the Bayes Filter Algorithm modified from the model in~\cite{Thrun} to reflect the two hidden states.

% \begin{enumerate}

%     \item Prior $\overline{bel}(s_{o}).$
    
%     \item $bel(s_{o}) = \eta P(z_{-i}(0) \: | \:  \overline{bel} (s_{o}))$
    
%     \item $t = 0$
    
%     \item Repeat until $t =$ length of observations:
    
%     For all $s_{t}$:
    
%     \tab \tab  $\overline{bel}(\hat{s_{t}})  =  \sum_{s_t}  P(\hat{s_{t}} \: | \: z_{i}(t),z_{-i}(t), s_{t})  bel(s_{t})$
    
%     \tab \tab $bel(\hat{s_{t}}) = \eta P (a_{-i}(t) \: | \: s_{t})  \overline{bel}(\hat{s_{t}})$

%     \tab For all $s_{t}$:
    
%     \tab \tab  $\overline{bel}(s_{t + 1}) = \sum_{\hat{s}_t}P(\hat{s_{t + 1}} \: | \: a_{i}(t),a_{-i}(t), \hat{s_{t}}) bel(\hat{s_{t}})$
    
%     \tab \tab $bel(s_{t + 1} ) = \eta P (z_{-i}(t) \: | \: s_{t + 1})  \overline{bel}({s_{t + 1}})$
    
%     \tab $t = t + 1$
% \end{enumerate}  

\begin{algorithm}
    \SetKwInOut{Input}{Input}
    \SetKwInOut{Output}{Output}
    \underline{function Bayes Filter} $( )$\;
    %\Input{list $goals$}
    %\Output{$goal$}
    $\overline{bel}(s_{0})$ \;
    $bel(s_{0}) = \eta P(z_{-i}(0) \: | \:  \overline{bel} (s_{0}))$\;
    \For{$t$ \KwTo $len(observations)$}
    {
     \For{$s_t \in states$}
    {
     $\overline{bel}(\hat{s_{t}})  =  \sum_{s_t}  P(\hat{s_{t}} \: | \: z_{i}(t),z_{-i}(t), s_{t})  bel(s_{t})$\; 
     $bel(\hat{s_{t}}) = \eta P (a_{-i}(t) \: | \: s_{t})  \overline{bel}(\hat{s_{t}})$\;
     }
     \For{$s_t \in states$}
    {
    $\overline{bel}(s_{t + 1}) = \sum_{\hat{s}_t}P({s_{t + 1}} \: | \: a_{i}(t),a_{-i}(t), \hat{s_{t}}) bel(\hat{s_{t}})$\;
    $bel(s_{t + 1} ) = \eta P (z_{-i}(t) \: | \: s_{t + 1})  \overline{bel}({s_{t + 1}})$\;
    } 
    } 
    \caption{\textbf{Bayes Filter Algorithm} }
    \label{alg:Bayes}
\end{algorithm}

\subsection{Predicting Actions, Plans, and Intents}
The Bayesian model and the Bayes Filter algorithm yield a probability distribution over hidden states in the model. From this distribution, one of the main tasks in this paper is to predict actions, plans, and intents. For each prediction, we will explore two methods: a MAP estimate and a more complete aggregation method. This subsection addresses how actions, plans, and intent can be predicted.

\subsubsection{Predicting an Action}
The results presented in this paper have been obtained for the Prisoners Dilemma game, so the set of possible actions contains {Cooperate, Defect}.
\paragraph{MAP Estimate for Action Prediction}
The MAP estimate takes the maximum of all the probabilities over the actions available to predict an action, $\hat{a}_{\rm MAP}$. The action probabilities are calculated by aggregating over all states as:

%For each action 'a':
%\tab $P(a) = \sum_{s \epsilon S} P(s) P(a |s)$
\[\hat{a}_{\rm MAP} = \arg \max_{a}\sum_{s \epsilon S} P(s) P(a |s)\]
%\newline The action with the highest probability is chosen as the prediction.

\paragraph {Aggregation Method for Action Prediction}
%Each expert $\phi_j$ has different states $s_{\phi_{j\_m}}$ with the probability distribution $P(s_{\phi_{j\_m}})$. 
%Each expert $\phi_i$ has different states $s \epsilon  S_{\phi_j} $ with the probability distribution $P(s)$. The 
Each expert $\phi_j$ has different states $s\in S_{\phi_j} $ with the probability distribution over each of these states $P(s_{\phi_j})$ generated by the Bayesian model. %The summation of the probability values for 
Summing probabilities for all the states that belong to a given expert is done for each expert giving $p(\phi_j) = \sum_{s\in S_{\phi_j}} P(s)$. The expert with the maximum probability is identified, and the action is selected with the equation below: 
%\tab $P(a) = \sum_{s \epsilon S_k} P(s) P(a |s)$
\[\hat{a}_{\phi} = \arg \max_{a}\sum_{s \epsilon S_{\phi_j}} P(s) P(a |s)\]

%action that would be chosen by this expert is used as the predicted action. This could be achieved using the formula:

%Predicted action for $\phi_j$= Target solution of $\arg \max_{\phi_j} \sum_{s\in S_{\phi_j}} P(s) $.
%\mg{I think it's better to let the $s$ in the summation be simple, i.e., $P(s)$, and let the index of summation indicate what values of $s$ are allowed, $\sum_{s\in S_{\phi_j}} $.}

\subsubsection{Predicting a Plan}
%\subsubsection{MAP Estimate of Plan}
%In the first approach, each expert used in S\# uses a state machine that maps action and speech act histories into the next action. Identifying the most likely expert predicts the plan of that expert. Thus, by associating each expert with a plan, the plan can be predicted by finding the MAP estimate over experts:
%\[
%    {\rm plan} = \arg \max_{\phi_i} \sum_{s\in S_{\phi_i}} P(s).
%\]

%\subsubsection{Aggregation Method for Plan Prediction}
%The second way to think about a plan is to group experts together by the strategy they use. 
We can categorize each expert as having a \textit{follower type} or a \textit{leader type}, as per the categorization of plan in~\cite{littman2001leading}.
A ``leader type'' creates strategies that will influence its partner to play a particular action by playing a trigger strategy that induces its partner to comply or be punished. Trigger strategies are the ones where a player begins by cooperation but defects for a predefined period of time when the other player shows a certain level of defection (opponent triggers through defection). The experts have a punishment stage in their state diagrams. The punishment phase is the strategy designed to minimize the partner’s maximum expected payoff. The punishment phase persists from the time the partner deviates from the offer until the sum of its partner’s payoffs (from the time of the deviation) is below what it would have obtained had it not deviated from the offer \cite{crandall2018cooperating}. Hence the partner's optimal strategy would be to follow the offer.
\iffalse
It includes the following experts generated by S\# for the Prisoners Dilemma game: 
    \begin{enumerate}
        \item Bouncer
        \item Leader pure- AX
        \item Leader alternating AY-BX
        \item Leader alternating AY-AX
        \item Leader alternating AX-BX
    \end{enumerate}
\fi
%like to bully, to cooperate, to protect, etc. 

A ``follower type'' expects its partner to do something and it plays the best response to their move. A follower assumes that its partner is using a trigger strategy. That means it assumes that its partner will propose an offer which is expected to be followed or else it might be punished. Following an offer may require the player to play fair (both getting the same payoff), to bully (demanding higher payoff than its associate) or to be bullied (accept lower payoffs than its associate). 
\iffalse
It includes the following experts generated by S\# for the Prisoners Dilemma game: 
\begin{enumerate}
        \item Maximin%- The Maximin strategy expects the other person to attack it, and the plan is to protect oneself. 
        \item Best Response%- The BR agent models the other player statistically and tries to create a model and try to maximize payoff with respect to that model.
        \item Follower- pure
        \item Follower alternating AY-BX
        \item Follower alternating AY-AX
        \item Follower alternating AX-BX
    \end{enumerate}
\fi

Two approaches can be used to estimate the leader/follower plan being used: MAP and Aggregation.

\paragraph{MAP Estimate for Plan Prediction}
Let $\theta(\phi_i)\in\{\mbox{ leader, follower}\}$ indicate the ``type'' of expert~$\phi_i$. The plan is then the most probable type. For this we first identify the MAP estimate for which expert is most likely and then select that expert's type,
\[
    \hat{\theta} = \theta\Big(\arg \max_{\phi_i} \sum_{s\in S_{\phi_i}} P(s) \Big).
\]

\paragraph{Aggregation Method for Plan Prediction}
Similar to how the probabilities of actions could be aggregated across states, we can aggregate probabilities across plan types and then choose the most likely type as follows:
\[
    \hat{\theta} = \arg \max_{{\rm type} \in \{\mbox{lead, foll}\}} \sum_{\theta(\phi_i)=={\rm type}}\sum_{s\in S_{\phi_i}} P(s).
\]

\subsubsection{Predicting Intent}
Each expert can be categorized by the goal it seeks to achieve by adopting its strategy. This is similar to categorizing agents by plan type, but the categorization is by intent type. 
We identify the intent using the Bayesian model of S\# using the simple rule: the intent of S\# is the intent of the expert it uses to achieve its goal. S\# experts fall into two goal types: ``Maximizing Payoff'' and ``Maximizing Fairness'' (by minimizing the difference between the two player's payoffs).

Intent can be predicted by identifying the intent type of the most likely expert using the MAP estimate, or by aggregating probabilities over intent types and then selecting the most probable type. These two prediction methods are exactly analogous to predicting plan type from the previous subsection so details are omitted.

\section{Experiments, Results and Discussion}

\subsection{Data preparation}
The dataset used in this work is from previous work by Crandall et al. (2018) \nocite{crandall2018cooperating} on RGs with cheap talk. Interaction logs have been recorded for human-human interaction and human interaction with S\#. Two players play Prisoners Dilemma against each other, each game lasting 51 rounds. For each round, each player gets an opportunity to share messages before taking their actions (which could include their plan to play a particular action, or anything they would want to say to their opponent). Interaction logs are formed based on those game logs, consisting of payoffs, cheap talk and actions played by the players in each gameplay. There are a total of 24 interactions, 12 human-human games and 12 human-S\# games, lasting 51~rounds each. 
%Since the interaction log files were in a text format, and contained many unwanted information, they needed to be processed to be transformed to necessary csv format, where each log file consists of the round number, proposals shared, and actions by the respective players.

Another dataset is formed by having the strategies shown in Table~\ref{tab:strat} play against both the S\# and human players. This dataset is used to compare the predictions of the proposed graphical Bayesian Model to evaluate its performance.

\subsection{Predicting Intent and Plans}
Two approaches were used for predicting the intent and the plan of the players for the repeated Prisoners Dilemma game: the MAP and aggregation method. %For each interaction, the intent of the players corresponds to the intent of the expert active at the time.
In our experiments, both the methods for predicting intent predicted ``Maximize payoff” as the intent of the players for all interactions. For predictions for S\#, the models' predictions comply with the actual intent of the experts of S\#. This is because the experts of S\# were designed in~\cite{crandall2018cooperating} with the intent to Maximize Payoff, except the Bouncer strategy which is never initialized for the Prisoners Dilemma game (Bouncer is relevant for other repeated games). 

For validating the plan prediction, the interaction history was run through S\# to see which of the experts were selected during each interaction, and hence the corresponding plan followed by the expert was considered as the true plan followed. 

When used for humans, the model also predicts the intent to be ``Maximizing Payoff". Unfortunately, we do not have measures to evaluate the intent prediction for humans for this game, which could be considered one of the limitations in our work. The intent prediction is based on the intent of the experts of S\#. It would have been interesting to evaluate the intent of the players from a different perspective like with respect to their personalities or motivational orientation as in the work~\cite{kuhlman1975individual}, where the goal of cooperative, competitive, and individualistic agents is to achieve joint gain, relative gain, and own gain respectively.

Both MAP and Aggregation approaches achieved an accuracy of $\approx$88\% for predicting plans for S\#. Paired T-test shows that the difference in average performance between the MAP approach and Aggregation for plan prediction is not big enough to be statistically significant (p = 0.0997). But for predictions without cheap talk, paired T-test show that the difference in average performance between the MAP approach and Aggregation is statistically significant (p = 0.0328), MAP being better. 

%Since only those S# experts are initiated for all rounds whose intent is to “Maximize payoff”, our model successfully predicted the intent of S#.  

\subsection{Predicting actions}

\paragraph{Average accuracy for action predictions}

\begin{figure}
    \centering
    \includegraphics[width=3.25in]{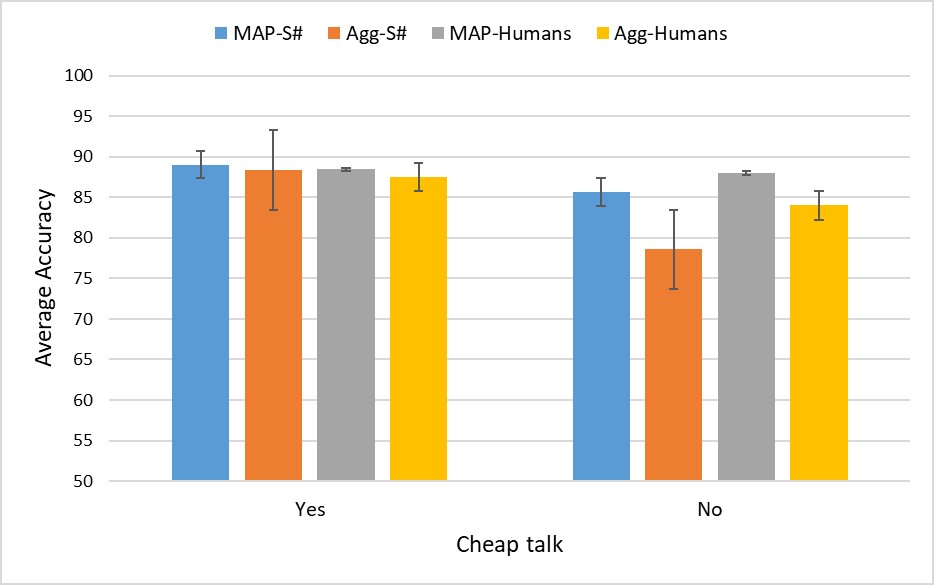}
    
    \caption{Average action prediction accuracy.}
    \label{fig:Average accuracy for action predictions}
\end{figure}
The average accuracy for predicting actions using the MAP and Aggregation was calculated for modeling S\#. Considering humans have similar internal states, the ability to form intentions, and plans to take actions, the same model was then used to model humans. 
%The Bayes Filter utilized the speech acts and previous actions as observations to find the posterior probabilities for each state transition. 
MAP performed better than Aggregation and was able to predict the actions 89.05\% of the time for S\#, and 88.45\% of the time for humans. We also tried experimenting on predicting the actions without using cheap talk and achieved an accuracy of 85.62\% for S\# and 88.02\% for humans. Figure~\ref{fig:Average accuracy for action predictions} summarizes the results.
Paired T-test shows that the difference in average performance between the MAP and Aggregation approaches is not big enough to be statistically significant (p = 0.338). However, without cheap talk, the paired T-test shows that MAP performs significantly better than Aggregation for predicting actions(p = 0.0328).
% But if the speech act signals are not present, paired T-test show that the difference in average performance between the MAP approach and Aggregation is statistically significant (p = 0.0328).

\iffalse
\begin{table}[htb]
    \begin{tabular}{ |p{1.8cm}|p{1.8cm} | p{1.8cm}|  }
     \hline
     \textbf{Model} & \textbf{Model S\#} & \textbf{Model Humans} \\ 
     \hline\hline
     MAP with Speech Acts & 89.05 & 88.45\\
     \hline
     Aggregation with Speech Acts & 88.40 & 87.53\\
     \hline
     MAP without Speech Acts & 85.62 & 88.02\\
     \hline
     Aggregation without Speech Acts & 78.59 & 83.99\\
     \hline
    \end{tabular}
\caption{Action prediction accuracy comparison for the two models}
\label{tab:map-agg}
\end{table}
\fi

\begin{figure}
    \centering
    \includegraphics[width=3.25in]{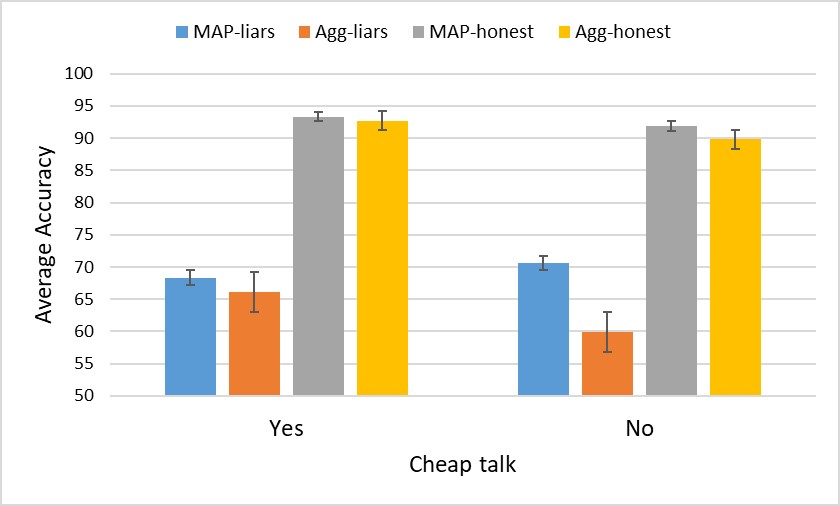}
    
    \caption{Action prediction comparison for players who lie.}
    \label{fig:comp-who-lie}
\end{figure}
\begin{figure}
    \centering
    \includegraphics[width=3.25in]{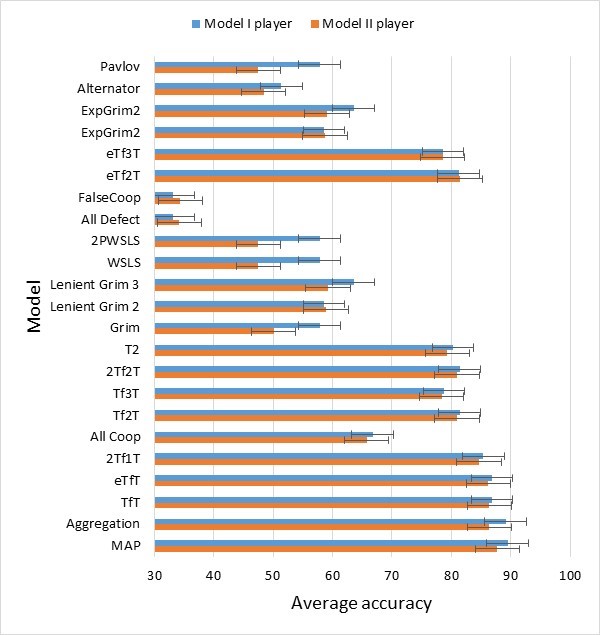}
    \caption{Comparison of action predictions for modeling Player 1 and 2 (Our model vs  Others).}
    \label{fig:comp1n2}
\end{figure}

There were 7 predictions where the accuracy was less than 80\%. Looking at the data, we found that this was because of players who lie ($\approx$51\% of the time on average). Lying refers to proposing a particular action but taking a different one during his/her turn. If we omit such interactions, MAP was 93.31\% accurate for modeling humans who lie less frequently (18.6\% of the time on average), and when we ignored the cheap talk, it was 92.2\% accurate. 

We see that the predictions were always better with cheap talk(without lying) as it provided more information about the interaction. It was interesting to see that the only time the accuracy bumped up when not using cheap talk, was when we modeled humans who lie. In this case, the accuracy of our model increased from 68.35\% to 70.59\% without cheap talk (for MAP approach). Thus, with this observation, we realize that our model performs well for modeling both S\# and humans except for the agents who lie. %Since S\# does not have an expert that takes care of the scenario of agents lying, a future enhancement for our model would be by creating such an expert for S\#. 
The results are presented in Figure~\ref{fig:comp-who-lie}.

\iffalse
\begin{table}[htb]
    \begin{tabular}{ |p{1.8cm}|p{1.8cm} | p{1.8cm}|  }
     \hline
     \textbf{Model} & \textbf{Model Humans who lie} & \textbf{Model Humans who lie less frequently} \\ 
     \hline\hline
     MAP with cheap talk & 68.35 & 93.31\\
     \hline
     Aggregation with cheap talk & 66.11 & 92.7\\
     \hline
     MAP without cheap talk & 70.59 & 92.2\\
     \hline
     Aggregation without cheap talk & 59.94 & 89.79\\
     \hline
    \end{tabular}
\caption{Action prediction comparison for the two models with and without players who lie}
\label{tab:map-agg-lie}
\end{table}
\fi
\subsection{Comparing MAP predictions to fixed models}

The action predictions from our model were compared with predictions from the fixed models presented in \cite{fudenberg2012slow}. The performance of the MAP and aggregation were comparable.  %Since, the Bayes Filter utilizes cheap talk and previous actions as observations to find the posterior probabilities for each state transition that determine which actions to take, unlike the fixed actions governed by the strategies in the literature, the performance based on our Bayesian model did the best out of other strategies. 
The Bayesian model outperformed the fixed strategies. However, it was interesting to see that Tit for tat performed nearly as well as our model for the action predictions. Also, the Exploitive Tit for Tat also performed very close to that of Tit for Tat.
%The other best strategy which did good(~85\%) was Two-tit-for-1-tat.

Figure~\ref{fig:comp1n2} shows how our model performed in modeling players 1 and 2 vs other strategies. The player number simply indicates the player who goes first in each game. The following subsection presents how our model performs better in modeling dynamic behavior in agents as compared to the fixed strategy models.

%\begin{figure}[htb]
%   \centering
%    \includegraphics[width=5in]{figs/comparison_of_MAP_action_predictions_1.jpg}
%    \caption{Comparison of action predictions for modeling Player 1 (Our model vs  Others)}
%    \label{fig:Comparison of action predictions for modeling Player 1 (Our model vs  Others)}
%\end{figure}
%\begin{figure}[htb]
%    \centering
%    \includegraphics[width=5in]{figs/comparison_of_MAP_action_predictions_2.jpg}
%    \caption{Comparison of action predictions for modeling Player 2 (Our model vs  Others)}
%    \label{fig:Comparison of action predictions for modeling Player 2 (Our model vs  Others)}
%\end{figure}

\subsection{Comparing MAP predictions excluding consistent interactions}
\begin{figure}
    \centering
    \includegraphics[width=3.25in]{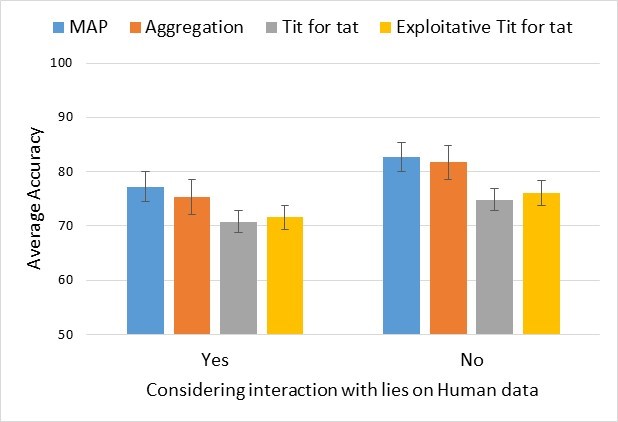}
    \caption{Performance with $<=$25\% continuous repetition in actions.}
    \label{fig:consistent-actions}
\end{figure}

Each of the interactions between all the players were observed carefully. Out of 48, 24 of the players had taken the same action repeatedly for more than 75\% of the rounds. For fixed strategies like Tit for tat, it is easier to make correct predictions in such cases, as interactions were consistent for at least 75\% of the rounds. So, another analysis was performed where we considered only those interactions which had more variance in its actions in each of the rounds. More specifically, interactions with only 25\% or fewer continuous repetition of the same action were taken. The results are presented in Figure~\ref{fig:consistent-actions}. Previously, we observed that Tit for tat  and exploitative Tit for tat performed very close to our models. But as we compare the interactions with more variance in actions across the rounds, our models were able to perform significantly better as given by paired T-test (p = 0.0160). In addition, as we disregarded the interactions including lies, the performance of MAP was increased to $\approx$82\%, and that of Aggregation increased to $\approx$81\%.

\iffalse
\begin{table}[htb]
    \begin{tabular}{ |p{1.8cm}|p{1.8cm} | p{1.8cm}|  }
     \hline
     \textbf{Model} & \textbf{Considering interactions with lies} & \textbf{Ignoring interactions with lies} \\ 
     \hline\hline
     MAP & 77.25 & 82.68\\
     \hline
     Aggregation & 75.29 & 81.70\\
     \hline
     Tit for tat & 70.78 & 74.84\\
     \hline
     Expoitative Tit for tat & 71.57 & 76.14\\
     \hline
    \end{tabular}
\caption{Performance with interactions with $<=$25\% consistency in actions}
\label{tab:consistent actions}
\end{table}
\fi
%\begin{figure}[htb]
%    \centering
%    \includegraphics[width=5in]{figs/comparison_of_MAP_action_predictions_notconstistent_1.jpg}
%    \caption{Comparison for MAP action predictions(excluding consistent interactions) for modeling player 1}
%    \label{fig:Comparison for MAP action predictions(excluding consistent interactions) for modeling player 1}
%\end{figure}
%\begin{figure}[htb]
%    \centering
%    \includegraphics[width=5in]{figs/comparison_of_MAP_action_predictions_notconstistent_2.jpg}
%    \caption{Comparison for MAP action predictions(excluding consistent interactions) for modeling player 2}
%    \label{fig:Comparison for MAP action predictions(excluding consistent interactions) for modeling player 2}
%\end{figure}
\begin{table}[h]
    \begin{footnotesize}
    \begin{tabular}{ |p{2.46cm}  p{5.20cm}|  }
     \hline
    \textbf{Strategy} & \textbf{Description} \\ 
     \hline
     Always Cooperate & Always play C\\ 
     \hline
     Tit-for-Tat (TFT) & Play C unless partner played D last round\\ 
     \hline
    TF2T & C unless D played in both last 2 rounds\\ 
        \hline
    TF3T & C unless D played in both last 3 rounds\\ 
        \hline
    2-Tits-for-1-Tat & Play C unless partner played D in either of the last 2 rounds (2 rounds of punishment if partner plays D)\\ 
        \hline
    2-Tits-for-2-Tats & Play C unless partner played 2 consecutive Ds in the last 3 rounds (2 rounds of punishment if D played twice in a row)\\ 
        \hline
    T2 & Play C until either player plays D, then play D twice and return to C\\ 
        \hline
    Grim & Play C until either plays D, then play D \\%forever\\ 
        \hline
    Lenient Grim 2 & Play C until 2 consecutive rounds occur in which either played D, then play D\\% forever\\ 
        \hline
    Lenient Grim 3 & Play C until 3 consecutive rounds occur in which either played D, then play D \\%forever\\ 
        \hline
    Perfect TFT/Win-StayLose-Shift & Play C if both players chose the same move last round, otherwise play D\\ 
        \hline
    Perfect Tit-for-Tat with 2 rounds of punishment & Play C if both players played C in the last 2 rounds, both players played D in the last 2 rounds, or both players played D 2 rounds ago and C last round. Otherwise play D\\ 
        \hline
    Always Defect & Always play D\\ 
        \hline
    False cooperator & Play C in the first round, then D forever\\ 
        \hline
    Expl.~Tit-for-Tat & Play D in the first round, then play TFT\\ 
        \hline
    Expl.~Tit-for-2-Tats & Play D in the first round, then play TF2T\\ 
        \hline
    Expl.~Tit-for-3-Tats & Play D in the first round, then play TF3T\\ 
        \hline
    Expl.~Grim2 & Play D in the first round, then play Grim2\\ 
        \hline
    Expl.~Grim3 & Play D in the first round, then play Grim3\\ 
        \hline
    Alternator & DCDC ...\\  
        \hline
    Pavlov & Start with C, Always play C if partner does not play D\\
        \hline
    
    \end{tabular}
    \end{footnotesize}

\caption{Existing strategies for Prisoners Dilemma.}
\label{tab:strat}
\end{table}
%\fi

\section{Conclusion}
This paper presented a graphical generative Bayesian model that models the S\# algorithm for two-player repeated games. The highlight of the model is its ability to model the internal states of S\#, considering each observation to calculate the posterior state probabilities, which could also be used to model humans. The other benefit of using this kind of model is that it is game independent, so it could be used for any two-player repeated game. 

In comparison with other strategy models, the MAP approach on the Bayesian model performed the best in predicting actions for the Prisoners Dilemma game. It could better model the dynamic actions of players as compared to the other fixed strategy models. Also, for both plan and action prediction, MAP performed significantly better without cheap talk, i.e. when it is an ordinary repeated game.
Additionally, obtaining a high accuracy for plans and intent prediction using the different approaches based on the same model, we can say that this Graphical Bayesian Model shows promise for modeling agents in two-player repeated games.

However, the model had some limitations. It was not able to detect its partners lying in the game and hence did not perform very well in such situations. A future enhancement could include creating experts for S\# having the ability to deal with lies in the game. Also, further exploration of the intent of players, based on other dimensions is necessary. It would have been interesting to study intent from a different perspective such as based on the personality of the players, and how the intention of players change over time. 

\section{Acknowledgements}
This work was supported in part by the U.S. Office of Naval Research under Grant \#N00014-18-1-2503. All opinions, findings, conclusions, and recommendations expressed in this paper are those of the author and do not necessarily reflect the views of the Office of Naval Research.

%\iffalse
\pagebreak

\bibliographystyle {aaai}
\bibliography {main}

\begin{thebibliography}{}

\bibitem[\protect\citeauthoryear{Axelrod and
  Hamilton}{1981}]{axelrod1981evolution}
Axelrod, R., and Hamilton, W.~D.
\newblock 1981.
\newblock The evolution of cooperation.
\newblock {\em science} 211(4489):1390--1396.

\bibitem[\protect\citeauthoryear{Baker, Saxe, and
  Tenenbaum}{2011}]{baker2011bayesian}
Baker, C.; Saxe, R.; and Tenenbaum, J.
\newblock 2011.
\newblock Bayesian theory of mind: Modeling joint belief-desire attribution.
\newblock In {\em Proceedings of the annual meeting of the cognitive science
  society}, volume~33.

\bibitem[\protect\citeauthoryear{Banerjee and Sen}{2007}]{banerjee2007reaching}
Banerjee, D., and Sen, S.
\newblock 2007.
\newblock Reaching pareto-optimality in prisoner’s dilemma using conditional
  joint action learning.
\newblock {\em Autonomous Agents and Multi-Agent Systems} 15(1):91--108.

\bibitem[\protect\citeauthoryear{Bratman}{1987}]{bratman1987intention}
Bratman, M.
\newblock 1987.
\newblock {\em Intention, plans, and practical reason}, volume~10.
\newblock Harvard University Press Cambridge, MA.

\bibitem[\protect\citeauthoryear{Brown}{1951}]{brown1951iterative}
Brown, G.~W.
\newblock 1951.
\newblock Iterative solution of games by fictitious play.
\newblock {\em Activity analysis of production and allocation} 13(1):374--376.

\bibitem[\protect\citeauthoryear{Cheng, Lo, and
  Leskovec}{2017}]{cheng2017predicting}
Cheng, J.; Lo, C.; and Leskovec, J.
\newblock 2017.
\newblock Predicting intent using activity logs: How goal specificity and
  temporal range affect user behavior.
\newblock In {\em Proceedings of the 26th International Conference on World
  Wide Web Companion},  593--601.
\newblock International World Wide Web Conferences Steering Committee.

\bibitem[\protect\citeauthoryear{Crandall \bgroup et al\mbox.\egroup
  }{2018}]{crandall2018cooperating}
Crandall, J.~W.; Oudah, M.; Ishowo-Oloko, F.; Abdallah, S.; Bonnefon, J.-F.;
  Cebrian, M.; Shariff, A.; Goodrich, M.~A.; Rahwan, I.; et~al.
\newblock 2018.
\newblock Cooperating with machines.
\newblock {\em Nature communications} 9(1):233.

\bibitem[\protect\citeauthoryear{Crandall}{2014}]{crandall2014towards}
Crandall, J.~W.
\newblock 2014.
\newblock Towards minimizing disappointment in repeated games.
\newblock {\em Journal of Artificial Intelligence Research} 49:111--142.

\bibitem[\protect\citeauthoryear{de Graaf and Malle}{2017}]{de2017people}
de~Graaf, M., and Malle, B.
\newblock 2017.
\newblock How people explain action (and ais should too).
\newblock In {\em Proceedings of the Artificial Intelligence for Human-Robot
  Interaction (AI-for-HRI) fall symposium}.

\bibitem[\protect\citeauthoryear{Deng and Deng}{2015}]{deng2015study}
Deng, X., and Deng, J.
\newblock 2015.
\newblock A study of prisoner’s dilemma game model with incomplete
  information.
\newblock {\em Mathematical Problems in Engineering} 2015.

\bibitem[\protect\citeauthoryear{Fudenberg, Rand, and
  Dreber}{2012}]{fudenberg2012slow}
Fudenberg, D.; Rand, D.~G.; and Dreber, A.
\newblock 2012.
\newblock Slow to anger and fast to forgive: Cooperation in an uncertain world.
\newblock {\em American Economic Review} 102(2):720--49.

\bibitem[\protect\citeauthoryear{Gaudesi \bgroup et al\mbox.\egroup
  }{2014}]{gaudesi2014turan}
Gaudesi, M.; Piccolo, E.; Squillero, G.; and Tonda, A.
\newblock 2014.
\newblock Turan: evolving non-deterministic players for the iterated prisoner's
  dilemma.
\newblock In {\em 2014 IEEE Congress on Evolutionary Computation (CEC)},
  21--27.
\newblock IEEE.

\bibitem[\protect\citeauthoryear{Kitano \bgroup et al\mbox.\egroup
  }{1997}]{kitano1997robocup}
Kitano, H.; Tambe, M.; Stone, P.; Veloso, M.; Coradeschi, S.; Osawa, E.;
  Matsubara, H.; Noda, I.; and Asada, M.
\newblock 1997.
\newblock The robocup synthetic agent challenge 97.
\newblock In {\em Robot Soccer World Cup},  62--73.
\newblock Springer.

\bibitem[\protect\citeauthoryear{Kuhlman and
  Marshello}{1975}]{kuhlman1975individual}
Kuhlman, D.~M., and Marshello, A.~F.
\newblock 1975.
\newblock Individual differences in game motivation as moderators of
  preprogrammed strategy effects in prisoner's dilemma.
\newblock {\em Journal of personality and social psychology} 32(5):922.

\bibitem[\protect\citeauthoryear{Lasota \bgroup et al\mbox.\egroup
  }{2017}]{lasota2017survey}
Lasota, P.~A.; Fong, T.; Shah, J.~A.; et~al.
\newblock 2017.
\newblock A survey of methods for safe human-robot interaction.
\newblock {\em Foundations and Trends{\textregistered} in Robotics}
  5(4):261--349.

\bibitem[\protect\citeauthoryear{Littman and Stone}{2001}]{littman2001leading}
Littman, M.~L., and Stone, P.
\newblock 2001.
\newblock Leading best-response strategies in repeated games.
\newblock In {\em In Seventeenth Annual International Joint Conference on
  Artificial Intelligence Workshop on Economic Agents, Models, and Mechanisms}.
\newblock Citeseer.

\bibitem[\protect\citeauthoryear{Malle and Knobe}{1997}]{malle1997folk}
Malle, B.~F., and Knobe, J.
\newblock 1997.
\newblock The folk concept of intentionality.
\newblock {\em Journal of experimental social psychology} 33(2):101--121.

\bibitem[\protect\citeauthoryear{Mollaret \bgroup et al\mbox.\egroup
  }{2015}]{mollaret2015perceiving}
Mollaret, C.; Mekonnen, A.~A.; Ferran{\'e}, I.; Pinquier, J.; and Lerasle, F.
\newblock 2015.
\newblock Perceiving user's intention-for-interaction: A probabilistic
  multimodal data fusion scheme.
\newblock In {\em 2015 IEEE International Conference on Multimedia and Expo
  (ICME)},  1--6.
\newblock IEEE.

\bibitem[\protect\citeauthoryear{Oudah \bgroup et al\mbox.\egroup
  }{2018}]{oudah2018ai}
Oudah, M.; Rahwan, T.; Crandall, T.; and Crandall, J.~W.
\newblock 2018.
\newblock How ai wins friends and influences people in repeated games with
  cheap talk.
\newblock In {\em Thirty-Second AAAI Conference on Artificial Intelligence}.

\bibitem[\protect\citeauthoryear{Park and Kim}{2016}]{park2016active}
Park, H., and Kim, K.-J.
\newblock 2016.
\newblock Active player modeling in the iterated prisoner's dilemma.
\newblock {\em Computational intelligence and neuroscience} 2016:38.

\bibitem[\protect\citeauthoryear{Perner}{1991}]{perner1991understanding}
Perner, J.
\newblock 1991.
\newblock {\em Understanding the representational mind.}
\newblock The MIT Press.

\bibitem[\protect\citeauthoryear{P{\^\i}rvu \bgroup et al\mbox.\egroup
  }{2018}]{pirvu2018predicting}
P{\^\i}rvu, M.~C.; Anghel, A.; Borodescu, C.; and Constantin, A.
\newblock 2018.
\newblock Predicting user intent from search queries using both cnns and rnns.
\newblock {\em arXiv preprint arXiv:1812.07324}.

\bibitem[\protect\citeauthoryear{Qu \bgroup et al\mbox.\egroup
  }{2019}]{qu2019user}
Qu, C.; Yang, L.; Croft, W.~B.; Zhang, Y.; Trippas, J.~R.; and Qiu, M.
\newblock 2019.
\newblock User intent prediction in information-seeking conversations.
\newblock In {\em Proceedings of the 2019 Conference on Human Information
  Interaction and Retrieval},  25--33.
\newblock ACM.

\bibitem[\protect\citeauthoryear{Rabkina and
  Forbus}{2013}]{Rabkina2013analogical}
Rabkina, I., and Forbus, K.~D.
\newblock 2013.
\newblock Analogical reasoning for intent recognition and action prediction in
  multi-agent systems.

\bibitem[\protect\citeauthoryear{Rios-Martinez \bgroup et al\mbox.\egroup
  }{2012}]{rios2012intention}
Rios-Martinez, J.; Escobedo, A.; Spalanzani, A.; and Laugier, C.
\newblock 2012.
\newblock Intention driven human aware navigation for assisted mobility.

\bibitem[\protect\citeauthoryear{Sen and Arora}{1997}]{sen1997learning}
Sen, S., and Arora, N.
\newblock 1997.
\newblock Learning to take risks.
\newblock In {\em AAAI-97 Workshop on Multiagent Learning},  59--64.

\bibitem[\protect\citeauthoryear{Stone and Veloso}{2000}]{stone2000multiagent}
Stone, P., and Veloso, M.
\newblock 2000.
\newblock Multiagent systems: A survey from a machine learning perspective.
\newblock {\em Autonomous Robots} 8(3):345--383.

\bibitem[\protect\citeauthoryear{Tavakkoli \bgroup et al\mbox.\egroup
  }{2007}]{tavakkoli2007vision}
Tavakkoli, A.; Kelley, R.; King, C.; Nicolescu, M.; Nicolescu, M.; and Bebis,
  G.
\newblock 2007.
\newblock A vision-based architecture for intent recognition.
\newblock In {\em International Symposium on Visual Computing},  173--182.
\newblock Springer.

\bibitem[\protect\citeauthoryear{Thrun, Burgard, and Fox}{2005}]{Thrun}
Thrun, S.; Burgard, W.; and Fox, D.
\newblock 2005.
\newblock {\em Probabilistic robotics}.
\newblock MIT press.

\end{thebibliography}

\end{document}